\documentclass{article} 
\usepackage{iclr2027_conference,times}

\usepackage{amsmath,amsfonts,bm}

\def\eqref#1{equation~\ref{#1}}

\def\1{\bm{1}}

\DeclareMathAlphabet{\mathsfit}{\encodingdefault}{\sfdefault}{m}{sl}
\SetMathAlphabet{\mathsfit}{bold}{\encodingdefault}{\sfdefault}{bx}{n}

\usepackage{hyperref}
\usepackage{url}
\usepackage{algorithm}
\usepackage{xcolor}
\usepackage{algpseudocode}
\usepackage{amsmath}
\usepackage{amssymb}
\usepackage{bibentry}
\usepackage[inline]{enumitem}
\usepackage{multirow}
\usepackage{booktabs}
\usepackage{array}
\usepackage{arydshln}
\usepackage{newfloat}
\usepackage{listings}
\usepackage{graphicx}
\usepackage{subcaption}
\usepackage{wrapfig}

\newcommand{\heading}[1]{\vspace*{1mm}\noindent\textbf{#1.}}

\title{Experimental Experience Modeling for \\ Autonomous Research}

\author{Wenda Wei$^{1,2,3}$, Yingchen Zhang$^{1,2,3}$,
Ruqing Zhang$^{1,2,3}$\thanks{Corresponding authors.},
Jiafeng Guo$^{1,2,3}$\footnotemark[1], Daiting Shi$^{4}$, \\
\textbf{Xueqi Cheng}$^{1,2,3}$ \\
$^{1}$State Key Laboratory of AI Safety \\
$^{2}$Institute of Computing Technology, Chinese Academy of Sciences \\
$^{3}$University of Chinese Academy of Sciences \\
$^{4}$Baidu Inc. \\
\texttt{\{weiwenda25z, zhangyingchen23s, zhangruqing, guojiafeng,} \\
\texttt{cxq\}@ict.ac.cn} \\
\texttt{shidaiting01@baidu.com} \\
}

\iclrfinalcopy

\begin{document}

\maketitle

\begin{abstract}
Autonomous research agents can generate hypotheses and conduct experiments, but experimentation remains a major source of computational cost. 
A fundamental challenge is deciding which experiments are worth running, particularly when prior evidence is insufficient to resolve uncertainty. 
Yet current research agents lack a systematic way to leverage experimental experience when making such decisions. 
We introduce Experimental Experience Modeling (EEM), a framework for making informed experimental decisions by acquiring, reusing, and accumulating experimental experience. 
EEM extracts decision-relevant records from earlier experimental trajectories, distills them into reusable experience, and organizes them in an experience library. 
For a new experimental decision, EEM retrieves relevant historical experience and assesses whether it provides sufficient support for deciding whether a candidate direction warrants further investment. 
When historical experience is insufficient, EEM conducts a targeted, low-cost pilot experiment to acquire the missing decision-relevant experience on demand. 
It then combines this newly acquired experience with retrieved historical experience to determine whether the direction warrants full-scale evaluation, which requires substantial resources.
The resulting experimental outcomes are further distilled into reusable experience, allowing the library to continually grow through iterative accumulation. 
Experiments on autonomous research benchmarks show that EEM improves research performance while reducing model interaction overhead, demonstrating the value of reusing accumulated experience and acquiring additional experience only when needed.

\end{abstract}

\section{Introduction}
\label{sec:introduction}

Recent advances in large language models (LLMs) have enabled autonomous research agents to coordinate literature review, hypothesis generation, experimentation, and scientific writing \citep{lu2024ai,schmidgall2025agent}. Recent systems further improve research ideas and implementations through iterative experimentation and feedback \citep{yuan2025dolphin,jiang2025aide,yamada2025ai}. 
As these agents become increasingly capable of conducting research autonomously, experimentation has emerged as a major source of computational cost.
The key challenge is not only \emph{what} to experiment with, but also \emph{which experiments are worth running}. An agent should avoid expensive experiments when existing evidence already supports a direction, yet should not over-rely on limited or uncertain evidence. Thus, effective research requires deciding whether to exploit existing evidence or acquire additional evidence through further experimentation.

Human researchers naturally accumulate experimental experience that goes beyond isolated results, capturing what works under particular conditions, what fails and why, and how such evidence should influence subsequent choices \citep{box1976science,collins1974tea}. 
When prior experience is insufficient, researchers can acquire targeted evidence through low-cost pilot experiments before committing substantial resources to full-scale evaluation \citep{thabane2010tutorial}. 
The resulting evidence is then incorporated into subsequent decisions. 
Thus, experimental experience provides a bridge between \emph{what has been learned before} and \emph{what should be done next}.

Recent autonomous research agents have begun to learn from previous attempts through feedback, reflection, and execution traces. 
Reflexion uses feedback from prior trials to improve subsequent attempts, while Dolphin incorporates experimental findings into further idea generation \citep{shinn2023reflexion,yuan2025dolphin}. 
AIDE, The AI Scientist-v2, and AutoResearchClaw similarly iteratively refine candidate implementations or execution strategies based on previous attempts \citep{jiang2025aide,yamada2025ai,liu2026autoresearchclaw}. 
However, these approaches primarily treat prior information as task-specific feedback or execution signals, rather than as reusable experimental experience explicitly organized to support future experimental decisions.
In particular, they do not systematically distill past experimental outcomes into reusable knowledge about \emph{when} a finding is applicable, \emph{what} empirical relationship it supports, and \emph{how} it should influence subsequent resource investment.
Nor do they explicitly assess whether available historical experience is sufficient for the current decision and, when it is not, acquire only the missing decision-relevant experience through a targeted low-cost experiment.

This motivates a fundamental question:

\emph{How can experimental experience be systematically acquired, reused, and accumulated to guide experimental decisions in autonomous research?}

To answer this question, we introduce \textbf{Experimental Experience Modeling (EEM)}, an experience-driven framework for efficient experimental decision making.
The framework contains three connected components.
(i) \emph{Historical experience acquisition.}
EEM identifies decision-relevant experimental records from prior experimental attempts within the research process and distills them into reusable experience.
Each experience item captures the conditions under which it is applicable, the empirical finding supported by the experiment, and the implication of that finding for future experimental investment, while retaining the original experimental record as supporting evidence.
(ii) \emph{Experience-guided experimental decision making.}
For a candidate direction, EEM first retrieves applicable historical experience and assesses whether it is sufficient to support the current decision.
If it is sufficient, EEM directly decides whether the direction should be pursued, revised, or rejected.
If it is insufficient, EEM identifies the missing decision-relevant experience and designs a targeted, low-cost pilot to acquire it before committing full-scale resources.
The newly acquired experience is then combined with the retrieved historical experience to support the final experimental decision.
(iii) \emph{Experience accumulation.}
Experience obtained from both pilot and full-scale experiments is added back to the historical library, allowing experience acquired for the current decision to become reusable prior experience for subsequent decisions.

We evaluate EEM on the 25 research topics in ARC-Bench \citep{liu2026autoresearchclaw}.
Compared with AutoResearchClaw \citep{liu2026autoresearchclaw}, the strongest evaluated baseline, EEM improves the overall score by 13.1\% and result analysis by 31.7\%, while reducing average total token consumption by 11.4\%.
Ablation experiments support the effectiveness of EEM's design, while experiments across backbone models and repeated runs examine its applicability and performance consistency.
\section{Methodology}
\label{sec:method}

We introduce Experimental Experience Modeling (EEM), a framework for experience-supported experimental decision making in autonomous research.
EEM reuses relevant historical experience to assess whether a candidate experiment warrants further resource investment and, when existing experience is insufficient, acquires the missing decision-relevant experience through a targeted, low-cost pilot.
We first describe the end-to-end research setting and provide an overview of EEM's design (Section~\ref{sec:setting}), then present historical experience acquisition (Section~\ref{sec:extraction}), experience-guided experimental decision making (Section~\ref{sec:utilization}), and iterative experience accumulation (Section~\ref{sec:iteration}).

\subsection{Overview}
\label{sec:setting}

\begin{figure*}[t]
    \centering
    \includegraphics[width=\textwidth]{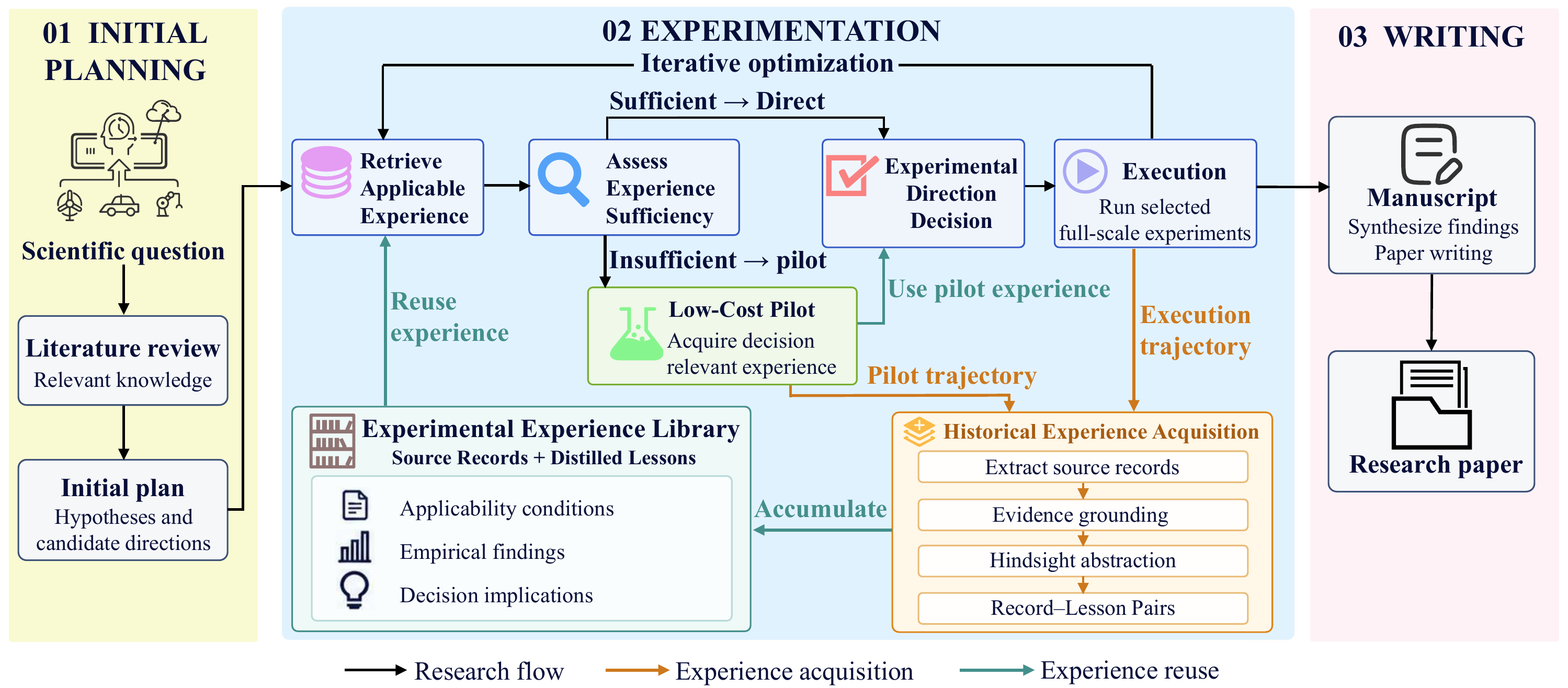}
    \vspace{-10pt}
    \caption{\textbf{Overview of Experimental Experience Modeling (EEM).}
    During the experimentation stage, EEM retrieves applicable experience to guide experimental decisions and conducts targeted, low-cost pilots only when existing experience is insufficient. Experience from both pilot and full-scale experiments is distilled into record--lesson pairs and accumulated for reuse in subsequent decisions within the research task.}
    \vspace{-10pt}
    \label{fig:framework}
\end{figure*}

We consider end-to-end autonomous research, where the input is a scientific question and the output is a research paper. As illustrated in Figure~\ref{fig:framework}, the workflow consists of three stages: (i) Initial planning reviews relevant literature and converts the research question into hypotheses and candidate experimental directions. (ii) Experimentation implements, evaluates, and iteratively refines these directions through empirical studies. (iii) Paper writing synthesizes the resulting motivation, methodology, experimental evidence, and conclusions into a research manuscript.

EEM focuses on the experimentation stage. Specifically, given a candidate direction under the current research state, EEM asks whether the available experimental experience is sufficient to determine whether the direction deserves further investment. It first retrieves relevant experience accumulated from previous experimental attempts. If this experience is sufficient, the direction can be directly pursued, revised, or rejected. Otherwise, EEM identifies the decision-relevant experience that is still missing and acquires it through a targeted pilot designed to obtain sufficient support at substantially lower cost than full-scale execution. The newly acquired experience is then combined with the retrieved historical experience to support the subsequent experimental decision, while completed experiments generate new reusable experience for later rounds. 
For each research task, EEM initializes the experience library as \(\mathcal{E}_0 = \varnothing\) and accumulates experience from experimental attempts conducted during the subsequent research process.

\subsection{Historical Experience Acquisition}
\label{sec:extraction}

EEM defines \emph{experimental experience} by its role in future experimental decisions: it should provide reusable, empirically grounded guidance for determining whether an experimental direction warrants further resource investment under a particular research context.
Accordingly, EEM first identifies decision-relevant records from previous experimental trajectories and then distills them into reusable experience.

\heading{Experience Definition and Representation}
A research trajectory contains many events that are not informative for future experimental decisions.
EEM therefore focuses on experimental records that connect a research context and an attempted direction to an observed consequence.
We represent each experience item as a source record paired with its distilled natural-language lesson:
\begin{equation}
    e_i = (r_i,\ell_i),
    \label{eq:experience}
\end{equation}
where $r_i$ records the experimental context, candidate direction, action, and observed outcome.
The lesson $\ell_i$ summarizes three components:
\begin{enumerate*}[label=(\roman*)]
    \item \emph{applicability conditions}, specifying the context and experimental conditions under which the finding holds and can guide a subsequent decision;
    \item \emph{empirical finding}, describing the observed relationship between the attempted direction and its outcome, bounded by the evidence available in the source record;
    \item \emph{decision implication}, expressing how the supported finding should guide future experimental choices and whether further resource investment is warranted.
\end{enumerate*}
The source record preserves the supporting evidence, ensuring that experience is derived from observed consequences rather than the agent's prior expectations.

Although all experience items share the representation in Eq.~\eqref{eq:experience}, they may capture different aspects of experimentation.
For lightweight organization, we therefore associate each item with one or more non-exclusive semantic tags, including \emph{mechanism effectiveness}, \emph{mechanism ineffectiveness and adverse effects}, \emph{errors and failure diagnoses}, \emph{experimental settings}, and \emph{experimental decisions}.
These tags provide an interpretable view of what an experience item concerns.

\heading{Extracting Experimental Experience}
EEM extracts experience from experimental attempts that produce consequences informative for future decisions.
Such records may originate from pilot experiments, full-scale evaluations, refinement attempts, or interrupted runs with informative partial results or failures.
Given a trajectory segment $\tau$, EEM first localizes decision-relevant records and then distills each record into an experience item:
\begin{equation}
\begin{aligned}
    \{r_i\}_{i=1}^{n_{\tau}}
    &= \operatorname{Record}_{\theta}(\tau), \\
    e_i
    &= \operatorname{Distill}_{\theta}(r_i), \\
    \Delta\mathcal{E}(\tau)
    &= \{e_i\}_{i=1}^{n_{\tau}},
\end{aligned}
\label{eq:extraction}
\end{equation}
where $\theta$ denotes the LLM parameters and $n_{\tau}$ is the number of localized records.
Each $\operatorname{Distill}_{\theta}(r_i)$ produces an experience item in the form of Eq.~\eqref{eq:experience}, pairing the source record with its distilled lesson.

The extraction process has two stages after record localization:
(i) \emph{evidence grounding} interprets the observed consequence with respect to the attempted direction, separating empirical support from the agent's original expectation.
(ii) \emph{hindsight abstraction} converts the grounded observation into reusable decision knowledge.
The distillation prompt therefore asks three questions: under what conditions the observation is applicable, what empirical relationship is supported by the record, and how that relationship should affect a future decision about further experimental investment.
In this way, EEM transforms a concrete observation such as ``a configuration improved a metric in this run'' into reusable guidance that states the applicable conditions, the supported finding about the direction, and its implication for future investment.

\subsection{Experience-Guided Experimental Decision Making}
\label{sec:utilization}

For each candidate experimental direction, EEM determines whether further resource investment is justified using the experience available at decision time.
The decision follows two possible paths:
(i) When relevant historical experience already provides sufficient support, EEM directly reuses that experience to determine whether the direction should be pursued, revised, or rejected.
(ii) When historical experience is insufficient, EEM identifies the missing decision-relevant experience, acquires it through a targeted, low-cost pilot, and then makes the decision using both historical and newly acquired experience.
In this way, EEM maximizes reuse of accumulated experience while acquiring additional experience only when necessary.

\heading{Decision Making with Sufficient Historical Experience}
For a candidate direction $d$ under research state $s_t$, EEM retrieves applicable historical experience as
$\mathcal{R}_t(d)=\operatorname{Retrieve}(\mathcal{E}_t;s_t,d,K)$,
where $\mathcal{E}_t$ is the historical experience library containing record--lesson pairs defined in Eq.~\eqref{eq:experience}, and $K$ denotes relevant model knowledge and literature-derived information.
The research state $s_t$ includes the current objective, experimental plan, and available results.

EEM then assesses whether the retrieved historical experience is sufficient for the current investment decision:
\begin{equation}
    (h_t(d),g_t(d))
    =
    \operatorname{Assess}_{\theta}
    \bigl(s_t,d,K,\mathcal{R}_t(d)\bigr),
    \label{eq:sufficiency}
\end{equation}
where $h_t(d)\in\{0,1\}$ indicates whether the available historical experience is sufficient, and $g_t(d)$ describes the decision-relevant experience gap when $h_t(d)=0$.
Experience sufficiency is decision-specific.
Historical experience may remain insufficient when it comes from mismatched conditions, contains inconsistent findings, or leaves an uncertainty critical to the current decision unresolved.

When $h_t(d)=1$, EEM directly makes the investment decision from the current state, relevant knowledge, and retrieved experience, choosing whether the direction should proceed to full-scale evaluation, be revised, or be rejected.
Applicable positive experience can justify further investment, while negative or failure-related experience can prevent repeated ineffective experimentation.

When $h_t(d)=0$, historical experience alone cannot reliably support the decision, and EEM follows the second path by acquiring the missing experience on demand.

\heading{Decision Making with On-Demand Experience Acquisition}
When historical experience is insufficient, EEM uses the identified gap $g_t(d)$ to determine what additional experience is required for the current decision.
EEM designs a targeted pilot that probes only the unresolved decision-critical uncertainty.
The pilot is not intended to approximate the entire full-scale experiment.
Its purpose is to obtain sufficient additional experience for the current decision at substantially lower cost.
EEM asks the model to specify the missing observation implied by $g_t(d)$ and construct a bounded pilot that isolates this observation while reducing experimental scale, duration, or coverage where possible.
After execution, the pilot trajectory $\tau_t^p(d)$ is processed by the same extraction procedure in Eq.~\eqref{eq:extraction}, yielding
$\Delta\mathcal{E}_t^p(d)=\operatorname{Extract}_{\theta}(\tau_t^p(d))$.
Each newly acquired item retains the record--lesson representation in Eq.~\eqref{eq:experience}.

The experience available for the current decision is therefore
\begin{equation}
    \mathcal{D}_t(d)=
    \begin{cases}
        \mathcal{R}_t(d),
        & h_t(d)=1,\\[3pt]
        \mathcal{R}_t(d)\cup\Delta\mathcal{E}_t^p(d),
        & h_t(d)=0.
    \end{cases}
    \label{eq:decision_experience}
\end{equation}
EEM then makes the final investment decision as
$a_t(d)=\operatorname{Decide}_{\theta}(s_t,d,K,\mathcal{D}_t(d))$,
where $a_t(d)$ specifies whether to proceed, revise, or reject the direction.
Historical experience provides reusable evidence accumulated from previous attempts, while the pilot supplies current, decision-targeted experience to address uncertainty specific to the present direction.
Their combination allows EEM to decide whether the direction should proceed to full-scale evaluation, be revised, or be abandoned before substantial resources are committed.

\subsection{Iterative Experience Accumulation}
\label{sec:iteration}

EEM closes the loop between experience reuse and experience acquisition by turning newly obtained experimental experience into historical experience for subsequent decisions.
After the decision and subsequent experimentation, the experience obtained from both pilot and full-scale execution is accumulated into the historical library and becomes available to later rounds.

Let $\Delta\mathcal{E}_t^p$ denote the union of the candidate-specific pilot experience sets $\Delta\mathcal{E}_t^p(d)$ acquired in round $t$.
Let $\tau_t^f$ denote the subsequent full-scale execution trajectory and
$\Delta\mathcal{E}_t^f=\operatorname{Extract}_{\theta}(\tau_t^f)$
the corresponding experience extracted using Eq.~\eqref{eq:extraction}.
The historical experience library is updated as
\begin{equation}
    \mathcal{E}_{t+1}
    =
    U\left(
        \mathcal{E}_t,
        \Delta\mathcal{E}_t^p
        \cup
        \Delta\mathcal{E}_t^f
    \right),
    \label{eq:iterative_update}
\end{equation}
where $U$ incorporates newly acquired record--lesson pairs into the library and removes duplicate entries corresponding to the same underlying decision record.
When no pilot is required, $\Delta\mathcal{E}_t^p=\varnothing$; when no full-scale experiment is conducted, $\Delta\mathcal{E}_t^f=\varnothing$.
When a candidate is not selected for full-scale execution, its pilot experience can still be retained because it provides empirical evidence relevant to future decisions.

In our setting, the experience library is initialized as $\mathcal{E}_0=\varnothing$ at the beginning of each research task.
Therefore, $\mathcal{E}_t$ contains experience accumulated from earlier pilot and full-scale experimental attempts within the same autonomous research run, and historical experience throughout this paper refers to this intra-task experimental history unless otherwise specified.
More generally, the same formulation allows $\mathcal{E}_0$ to be warm-started with experience collected from previous research tasks, provided that its applicability conditions are compatible with the current problem.
We leave systematic cross-task experience transfer and the associated risk of negative transfer to future work.

This accumulation process gradually converts local experimental outcomes into reusable historical experience.
As the library grows, later decisions can draw on a broader set of prior observations, reducing the need to reacquire experience that has already been established under applicable conditions.
The latest experimental results and accumulated experience then inform revisions to the plan and the candidate directions considered in the next round, closing the cycle of experience reuse, on-demand acquisition, and accumulation.
The experimental loop terminates when the collected evidence adequately addresses the research question and supports the intended claims, or when the configured iteration limit is reached.
EEM then proceeds to paper writing using the accumulated experimental record.

\section{Results}
\label{sec:results}

We evaluate EEM on ARC-Bench \citep{liu2026autoresearchclaw} to examine whether reusable experimental experience and on-demand pilot acquisition improve autonomous research. Our experiments focus on five aspects:
\begin{enumerate*}[label=(\roman*)]
\item overall performance: comparison with representative autonomous research systems across code development, code execution, and result analysis (Section~\ref{sec:overall_performance});
\item efficiency: comparison of model token consumption across planning, experimentation, and writing (Section~\ref{sec:efficiency});
\item component effectiveness: ablations of experience distillation, on-demand pilot acquisition, and historical experience reuse (Section~\ref{sec:ablation});
\item backbone applicability: evaluation with a general-purpose model in addition to a coding-oriented model (Section~\ref{sec:backbone}); and
\item stability across runs: repeated executions under the same configuration to assess performance consistency (Section~\ref{sec:stability}).
\end{enumerate*}

\subsection{Experimental Setup}
\label{sec:experimental_setup}

\heading{Benchmark and Evaluation Metrics}
We use ARC-Bench \citep{liu2026autoresearchclaw}, which contains 25 machine learning research topics with specified research objectives and experimental requirements.
The benchmark evaluates three aspects of experimentation: Code Development (CD), Code Execution (CE), and Result Analysis (RA).
CD assesses the implementation of the proposed methods and baselines, CE measures successful execution and the production of required experimental evidence, and RA evaluates whether the conclusions are supported by the observed results.
An LLM-based evaluator assigns scores according to the benchmark rubrics, with higher scores indicating better performance.
The overall score is computed as $0.25\,\mathrm{CD} + 0.25\,\mathrm{CE} + 0.50\,\mathrm{RA}$.

\heading{Baselines and Implementation}
We compare EEM with three autonomous research baselines: 
(i) AI Scientist v2 \citep{yamada2025ai} automates the research workflow and uses agentic tree search to improve experiments.
(ii) AIDE-ML \citep{jiang2025aide} searches over candidate implementations through iterative code generation, evaluation, and refinement.
(iii) AutoResearchClaw \citep{liu2026autoresearchclaw} coordinates an end-to-end research pipeline with adaptive experimentation and feedback; we use its fully autonomous version.
All baselines and EEM use the same GPT-5.3-Codex backbone and Python sandbox settings and are evaluated on identical benchmark tasks using the same scoring protocol.
We limit each pilot experiment to 120 seconds to prevent excessively long pilot runs.

\begin{table}[t]
\centering
\caption{Main experimental results of EEM and baselines on ARC-Bench. The overall score weights Code Development, Code Execution, and Result Analysis at 25:25:50. Bold values indicate the best results. Higher scores are better.}
\label{tab:main_results}
\begin{tabular}{lcccc}
\toprule
\textbf{Framework} & \textbf{Code Dev} & \textbf{Code Exec} &
\textbf{Result Analysis} & \textbf{Overall} \\
\midrule
AI Scientist v2              & 0.712 & 0.442 & 0.261 & 0.419 \\
AIDE-ML                      & \textbf{0.958} & 0.415 & 0.336 & 0.511 \\
AutoResearchClaw (Full-Auto) & 0.938 & 0.562 & 0.442 & 0.596 \\
\midrule
EEM & 0.911 & \textbf{0.619} & \textbf{0.582} & \textbf{0.674} \\
\bottomrule
\vspace{-5pt}
\end{tabular}
\end{table}

\subsection{Overall Performance}
\label{sec:overall_performance}

Table~\ref{tab:main_results} shows that EEM achieves the highest overall score of 0.674, compared with 0.596 for AutoResearchClaw, 0.511 for AIDE-ML, and 0.419 for AI Scientist v2.
The 13.1\% improvement over AutoResearchClaw, the strongest evaluated baseline, supports the effectiveness of EEM's experience-supported experimental decision framework.
EEM combines reusable historical experience with targeted pilot acquisition when additional decision-relevant experience is needed, providing a basis for deciding which directions warrant further investment.

EEM improves code execution from 0.562 to 0.619 and result analysis from 0.442 to 0.582, despite a lower code development score than AutoResearchClaw.
These results indicate that EEM's overall advantage lies in successfully conducting experiments and drawing supported conclusions. The largest improvement occurs in result analysis, with a relative gain of 31.7\%.
This metric evaluates whether conclusions are supported by observed results, which is aligned with EEM's emphasis on grounding reusable experience in concrete experimental records.

\subsection{Efficiency}
\label{sec:efficiency}

Table~\ref{tab:token_cost} compares the average token consumption of EEM and AutoResearchClaw across the 25 ARC-Bench topics.
Input tokens include the prompts and context supplied to the model, while output tokens include its generated responses.
We report both quantities for planning, experimentation, and writing.

\begin{wraptable}{r}{0.66\textwidth}
\vspace{-12pt}
\centering
\scriptsize
\renewcommand{\arraystretch}{1.08}
\setlength{\tabcolsep}{2.7pt}
\captionsetup{width=\linewidth}
\caption{Efficiency comparison of EEM and AutoResearchClaw in LLM input and output token consumption across research stages.}
\vspace{-8pt}
\label{tab:token_cost}
\resizebox{\linewidth}{!}{%
\begin{tabular}{@{}llrrrr@{}}
\toprule
\textbf{Method} & \textbf{Token Type}
& \textbf{Planning}
& \textbf{Experimentation}
& \textbf{Writing}
& \textbf{Total} \\
\midrule
\multirow{2}{*}{AutoResearchClaw}
& Input
& 53,581 & 265,000 & 227,612 & 546,193 \\
& Output
& \textbf{27,003} & 209,330 & \textbf{60,758} & 297,091 \\
\midrule
\multirow{2}{*}{EEM}
& Input
& \textbf{48,064} & \textbf{230,240}
& \textbf{220,353} & \textbf{498,657} \\
& Output
& 28,314 & \textbf{156,984}
& 62,902 & \textbf{248,200} \\
\bottomrule
\end{tabular}%
}
\vspace{-8pt}
\end{wraptable}

EEM reduces total token consumption from 843,284 to 746,857 tokens per topic, a reduction of 11.4\%.
Input and output tokens decrease by 8.7\% and 16.5\%, respectively.
Together with its higher overall score, these results show that EEM improves research performance with lower model interaction overhead, despite the additional operations required for experience distillation and pilot-based experience acquisition. The largest savings occur during experimentation, where combined token consumption decreases from 474,330 to 387,224, a reduction of 18.4\%.
This pattern is consistent with EEM's design principle of reusing historical experience when it is sufficient and acquiring additional experience only when necessary.

\subsection{Ablation Study}
\label{sec:ablation}

Table~\ref{tab:ablation} reports the results of three ablations aligned with the main components of EEM:
\begin{enumerate*}[label=(\roman*)]
    \item \textit{w/o experience distillation}, which retains historical source records but removes their distilled reusable lessons and associated retrieval metadata;
    \item \textit{w/o on-demand acquisition}, which disables pilot experiments so that no additional decision-targeted experience is acquired before the full-scale decision; and
    \item \textit{w/o historical experience}, which removes the accumulated experience library.
\end{enumerate*}
All variants are evaluated on ARC-Bench under the same experimental settings as the full framework, except for the removed component.

\begin{wraptable}{r}{0.66\textwidth}
\centering
\scriptsize
\renewcommand{\arraystretch}{1.08}
\setlength{\tabcolsep}{3.1pt}
\captionsetup{width=\linewidth}
\vspace{-5pt}
\caption{Ablation study of EEM. Each variant removes one component: (i) experience distillation, while retaining source records; (ii) on-demand pilot acquisition; or (iii) the historical experience library.}
\vspace{-5pt}
\label{tab:ablation}
\resizebox{\linewidth}{!}{%
\begin{tabular}{@{}lcccc@{}}
\toprule
\textbf{Variant} & \textbf{Code Dev} & \textbf{Code Exec} &
\textbf{Result Analysis} & \textbf{Overall} \\
\midrule
EEM & \textbf{0.911} & \textbf{0.619} &
\textbf{0.582} & \textbf{0.674} \\
w/o experience distillation & 0.874 & 0.598 & 0.563 & 0.649 \\
w/o on-demand acquisition & 0.798 & 0.454 & 0.504 & 0.565 \\
w/o historical experience & 0.671 & 0.413 & 0.470 & 0.506 \\
\bottomrule
\end{tabular}%
}
\vspace{-5pt}
\end{wraptable}

EEM achieves the best performance across all metrics.
The full framework obtains an overall score of 0.674, compared with 0.649 without experience distillation, 0.565 without on-demand acquisition, and 0.506 without historical experience.
These results provide end-to-end evidence that reusable historical experience and targeted acquisition of additional experience make complementary contributions to EEM's experimental decision process:
(i) Removing historical experience causes the largest performance decrease, lowering the overall score from 0.674 to 0.506, a relative decrease of 24.9\%.
This result highlights the importance of accumulating and reusing information from previous experimental attempts rather than repeatedly reasoning from the current state alone.
(ii) Removing on-demand acquisition also causes a substantial drop to 0.565, indicating that historical experience by itself does not fully replace the need to acquire additional evidence when the current decision remains unresolved.
(iii) Retaining source records without experience distillation yields 0.649, suggesting that transforming concrete experimental history into reusable decision-oriented guidance provides additional value beyond storing raw records alone.

\begin{figure}[t]
\centering
\begin{subfigure}[t]{0.48\columnwidth}
    \centering
    \includegraphics[width=\linewidth]{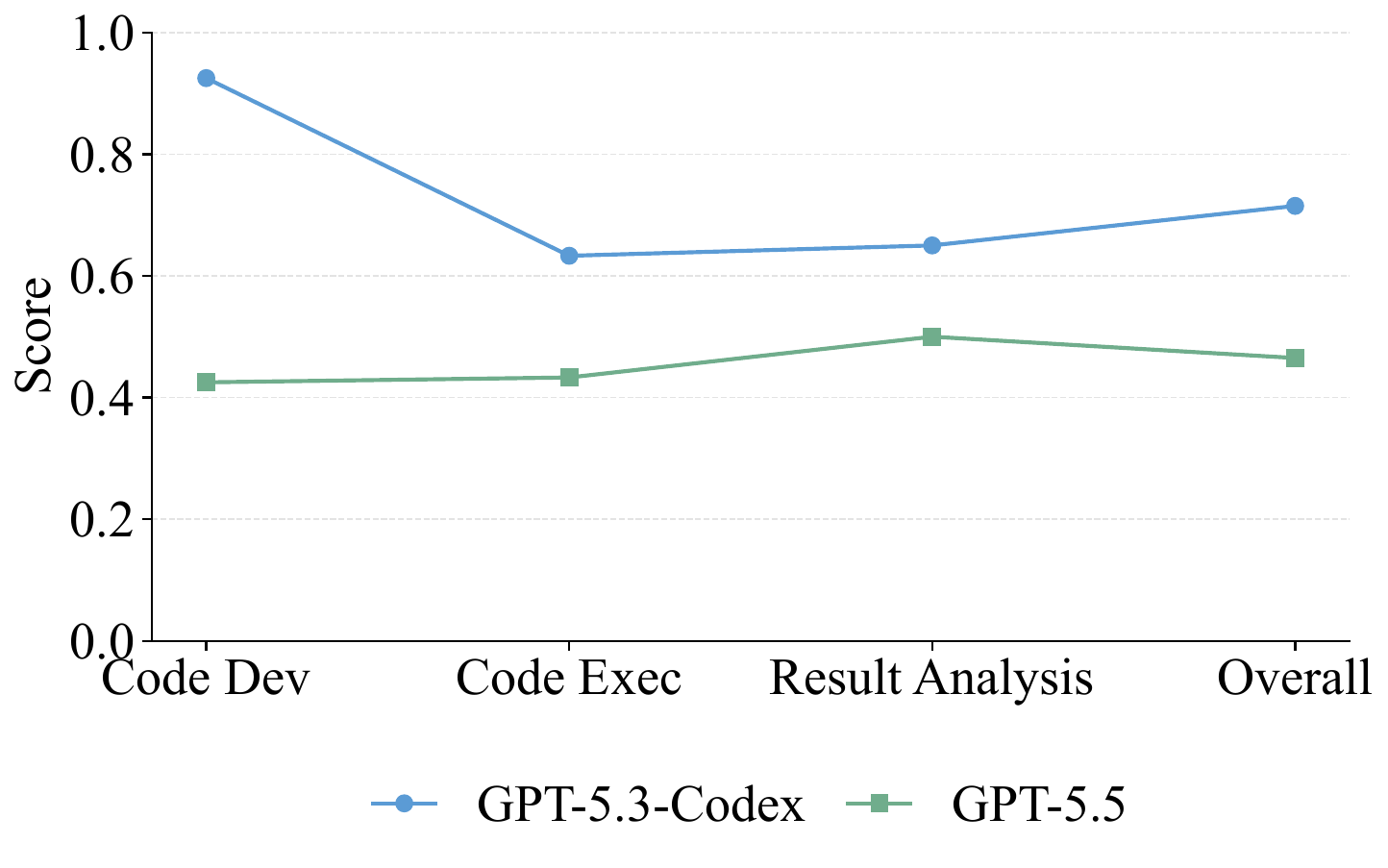}
    \caption{Performance with different backbone models.}
    \label{fig:run_results}
\end{subfigure}
\hfill
\begin{subfigure}[t]{0.48\columnwidth}
    \centering
    \includegraphics[width=\linewidth]{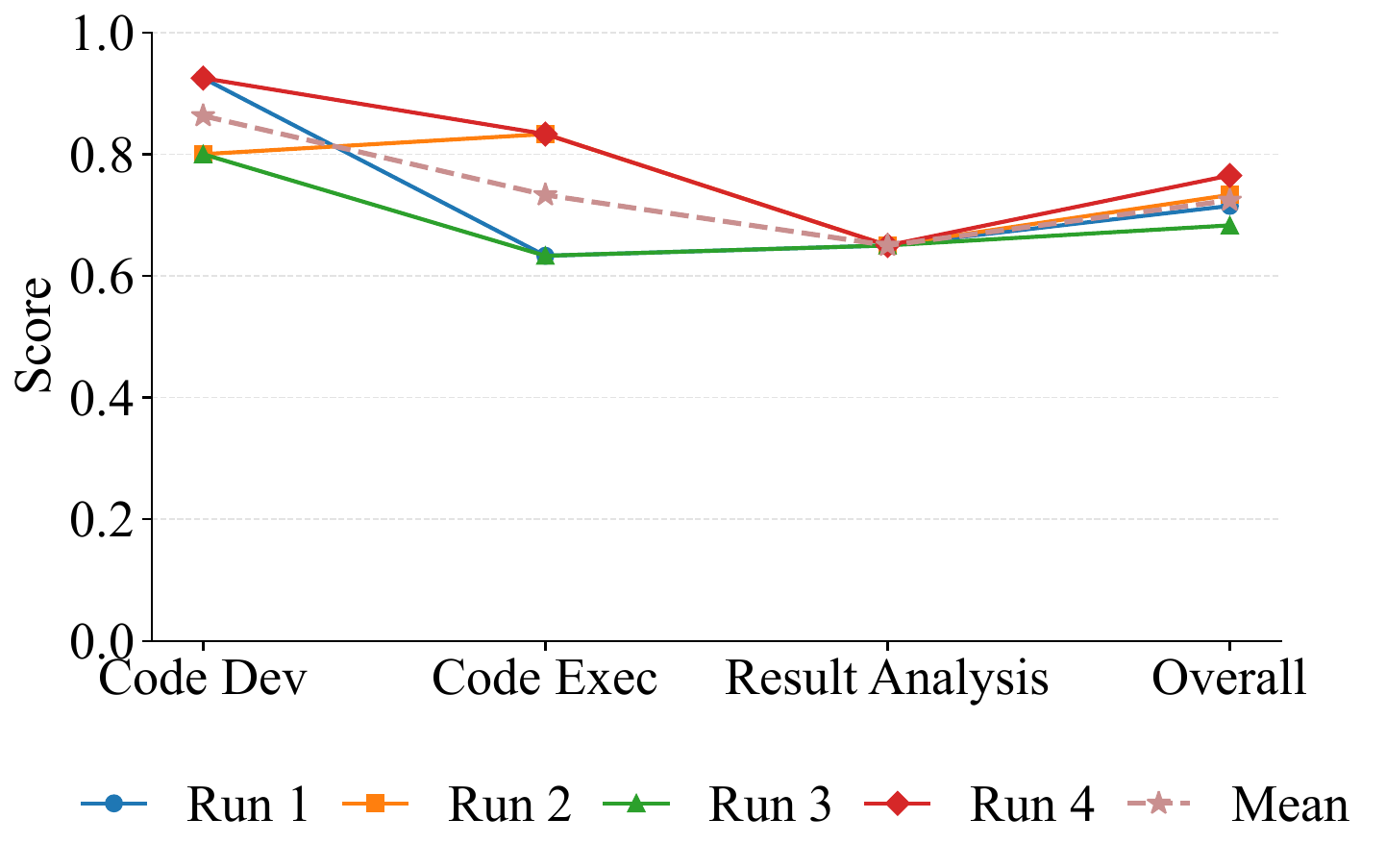}
    \caption{Stability across four independent runs.}
    \label{fig:token_results}
\end{subfigure}

\caption{Backbone applicability and stability analysis of EEM on ARC-Bench task ML10. (a) Performance using GPT-5.3-Codex and GPT-5.5 under the same framework configuration. (b) Performance across four independent runs with the same model and configuration.}
\vspace{-5pt}
\label{fig:analysis}
\end{figure}

\subsection{Applicability Across Backbone Models}
\label{sec:backbone}

Figure~\ref{fig:analysis}(a) compares EEM using GPT-5.3-Codex and GPT-5.5 on ARC-Bench task ML10 under the same framework configuration.
With GPT-5.5 as a general-purpose backbone, EEM completes the full workflow from a research question through experimentation to manuscript generation, showing that the framework can operate with both a coding-oriented and a general-purpose backbone in the evaluated task. GPT-5.5 obtains lower scores than GPT-5.3-Codex in this comparison, with the largest decrease in code development.
This suggests that implementation quality remains sensitive to the backbone's coding capabilities even when experimental decision making and experience reuse are provided by the framework.
The result-analysis score declines less than the code-development score, indicating that changing the backbone affects evaluation dimensions differently.

\subsection{Stability Across Runs}
\label{sec:stability}

Figure~\ref{fig:analysis}(b) reports four independent executions of EEM on ARC-Bench task ML10.
The overall scores remain within an approximately 0.68--0.77 range.
This supports repeatability under the evaluated configuration rather than dependence on a single successful execution.
Specifically, result analysis remains unchanged across all four runs, while code development and code execution vary without substantially affecting overall performance.
Overall, EEM exhibits limited variation in its aggregate score across these repeated runs.

\section{Related Work}
\label{sec:related_work}

\heading{Autonomous Research}
Autonomous research systems increasingly connect scientific reasoning with tools, experimentation, and manuscript preparation.
This progress is enabled by the growing capabilities of LLMs \citep{anthropic2026claudesonnet5,openai2026gpt53codex,openai2026gpt55}.
Early systems such as Coscientist and ChemCrow integrate LLMs with scientific tools to support chemical planning and experimentation~\citep{boiko2023autonomous,m2024augmenting}.
The AI Scientist, Agent Laboratory, and AI-Researcher extend this direction toward integrated workflows covering literature review, implementation, evaluation, and writing~\citep{lu2024ai,schmidgall2025agent,tang2026ai}.
Beyond workflow integration, Dolphin uses experimental feedback to refine research ideas, while AIDE and The AI Scientist-v2 explore and improve candidate implementations through search~\citep{yuan2025dolphin,jiang2025aide,yamada2025ai}.
The AI co-scientist further investigates iterative hypothesis generation through debate and tournament-based refinement~\citep{gottweis2025towards}.
AutoResearchClaw combines adaptive execution, hypothesis revision, and lessons from previous failures within an autonomous research pipeline~\citep{liu2026autoresearchclaw}.
Building on these advances, EEM focuses specifically on experimental decision making in autonomous research, reusing accumulated experimental experience and acquiring additional experience through targeted pilots only when needed.

\heading{Experience Reuse in Language Agents}
Experience reuse builds on the principle of adapting previous cases to new decisions~\citep{aamodt1994case}.
For language agents, Generative Agents, MemGPT, and A-MEM investigate complementary mechanisms for retaining, retrieving, and organizing information across interactions~\citep{park2023generative,packer2023memgpt,xu2026mem}.
Reflexion and ExpeL extract verbal reflections or reusable insights from prior attempts, while Voyager retains executable skills for subsequent tasks~\citep{shinn2023reflexion,zhao2024expel,wang2023voyager}.
More recent approaches develop persistent strategy memories: Dynamic Cheatsheet maintains an adaptive record of useful knowledge, ReasoningBank distills strategies from successful and failed trajectories, and ACE incrementally organizes experience into evolving playbooks~\citep{suzgun2026dynamic,ouyang2026reasoningbank,zhang2026agentic}.
Within autonomous research, AgentRxiv enables agents to share and build on previous research reports~\citep{schmidgall2025agentrxiv}.
Dream-RSI further reuses accumulated discovery history by constructing replay simulators from historical discovery trees, enabling exploration policies to be evaluated and improved without repeatedly invoking expensive online execution~\citep{zheng2026dream}.
EEM differs in focusing specifically on reusable \emph{experimental decision experience}.
Rather than retaining general interaction memory, research reports, or replaying historical search trajectories, EEM distills experimental records into experience that specifies its applicability conditions, empirically supported finding, and implication for future resource investment.
This experience is used directly to determine whether an experimental direction warrants further investment and whether additional experience must first be acquired.

\heading{Cost-Efficient Experimental Decision Making}
A related line of work reduces expensive evaluation through preliminary or lower-fidelity experiments. Pilot studies collect limited evidence before full-scale studies~\citep{thabane2010tutorial}, while successive halving, Hyperband, and multi-fidelity optimization use cheaper evaluations to allocate resources among candidate configurations~\citep{jamieson2016nonstochastic,li2018hyperband,kandasamy2017multifidelity}. These methods mainly study resource allocation under predefined candidates, objectives, or fidelity levels. EEM instead addresses open-ended autonomous research, where different experimental directions may require different evidence for deciding whether they are worth pursuing. Rather than routinely evaluating candidates at lower fidelity, EEM first reuses historical experience and invokes a targeted pilot only when the experience required for the current decision is missing.

Most closely related, AI Research Preference Models (RPMs) predict which research candidates are worth executing, with an agentic variant using small-scale pilots to improve candidate ranking~\citep{foster2026airesearchpreference}. EEM instead formulates the problem as experience-supported experimental decision making: it distills prior experiments into reusable experience, assesses whether that experience is sufficient, and acquires only the missing decision-relevant experience when necessary. The resulting experience supports not only candidate selection, but also decisions to pursue, revise, or reject an experimental direction.
\section{Conclusion}
\label{sec:conclusion}

We introduced Experimental Experience Modeling (EEM), a framework for experience-supported experimental decision making in autonomous research. 
EEM distills experimental trajectories into reusable experience, retrieves applicable experience to guide resource investment, and acquires missing decision-relevant experience through targeted, low-cost pilots when historical experience is insufficient. 
Experience from both pilot and full-scale experiments is accumulated for subsequent decisions. 
Experiments on ARC-Bench show that EEM improves research performance while reducing model token consumption, with ablation studies supporting the effectiveness of its design.
These results suggest that reusing accumulated experimental experience and selectively acquiring additional evidence can improve the effectiveness and efficiency of autonomous experimentation.
Future work will explore experience transfer across research tasks to support broader reuse of accumulated knowledge. 
We will also investigate how to better balance experience reuse and evidence acquisition in autonomous research.

\clearpage

\bibliography{iclr2027_conference}
\bibliographystyle{iclr2027_conference}


\end{document}